\documentclass[runningheads]{llncs}

\usepackage{eccv}

\usepackage{eccvabbrv}

\usepackage{graphicx}
\usepackage{booktabs}

\usepackage[accsupp]{axessibility}  

\usepackage[pagebackref,breaklinks,colorlinks,citecolor=eccvblue]{hyperref}

\usepackage{orcidlink}
\usepackage{adjustbox}
\usepackage{tabu}
\usepackage{verbatim}
\usepackage{xcolor, colortbl}
\usepackage{tcolorbox}
\usepackage{amsmath}
\usepackage{pifont}
\usepackage{multirow}
\newcommand\textvtt[1]{{\normalfont\fontfamily{cmvtt}\selectfont #1}}
\begin{document}

\title{Reflective Instruction Tuning: Mitigating Hallucinations in Large Vision-Language Models} 

\titlerunning{Reflective Instruction Tuning}

\author{Jinrui Zhang\inst{1}\orcidlink{0000-0001-5531-4160},
Teng Wang\inst{1,2}\orcidlink{0000-0003-2331-3619},
Haigang Zhang\inst{3}\orcidlink{0000-0002-8461-766X}, Ping Lu\inst{4}, Feng Zheng\inst{1,5}\orcidlink{0000-0002-1701-9141}\thanks{Corresponding author.}}

\authorrunning{J.~Zhang et al.}

\institute{Southern University of Science and Technology
\and
The University of Hong Kong\and Shenzhen Polytechnic University
\and
The Cloud Computing and IT Institute of ZTE Corporation\\
\and
Research Institute of Multiple Agents and Embodied Intelligence,\\ Peng Cheng Laboratory, Shenzhen, China
}

\maketitle

\begin{abstract}
Large vision-language models (LVLMs) have shown promising performance on a variety of vision-language tasks. However, they remain susceptible to hallucinations, generating outputs misaligned with visual content or instructions. While various mitigation strategies have been proposed, they often neglect a key contributor to hallucinations: lack of fine-grained reasoning supervision during training. Without intermediate reasoning steps, models may establish superficial shortcuts between instructions and responses, failing to internalize the inherent reasoning logic. To address this challenge, we propose reflective instruction tuning, which integrates rationale learning into visual instruction tuning. Unlike previous methods that learning from responses only, our approach entails the model predicting rationales justifying why responses are correct or incorrect. This fosters a deeper engagement with the fine-grained reasoning underlying each response, thus enhancing the model’s reasoning proficiency. To facilitate this approach, we propose \textit{REVERIE}, the first large-scale instruction-tuning dataset with \textit{ReflEctiVE RatIonalE} annotations. \textit{REVERIE} comprises 115k machine-generated reasoning instructions, each meticulously annotated with a corresponding pair of correct and confusing responses, alongside comprehensive rationales elucidating the justification behind the correctness or erroneousness of each response. Experimental results on multiple LVLM benchmarks reveal that reflective instruction tuning with the \textit{REVERIE} dataset yields noticeable performance gain over the baseline model, demonstrating the effectiveness of reflecting from the rationales. Project page is at \url{https://zjr2000.github.io/projects/reverie}

  \keywords{Large Vision Language Models \and Visual Instruction Tuning \and Hallucination Mitigation}
\end{abstract}

\section{Introduction}

Recently, large vision-language models (LVLMs)~\cite{zhu2023minigpt, liu2024visual, zhang2023llavar, liu2023improved, instructblip, ye2023mplug, chen2023shikra, bai2023qwen} have garnered significant interest within the research community. By leveraging large language models (LLMs)~\cite{touvron2023llama, zheng2023judging, touvron2023llama2, chowdhery2023palm, chung2022scaling, ouyang2022training} as a unified task interface and employing visual instruction tuning~\cite{liu2024visual} for enhanced task-level generalization, LVLMs have showcased remarkable reasoning performance and robustness in following user instructions. These breakthroughs offer substantial promise for the evolution of artificial general intelligence. Despite these advancements, LVLMs continue to be impeded by the phenomenon of hallucination~\cite{li2023evaluating, guan2023hallusionbench}, wherein the models generate outputs that are inconsistent with visual content, established world knowledge, or the instructions provided. This significantly undermines the reliability of such models in real-world scenarios. Extensive research has investigated the origins of hallucinations in LVLMs, identifying factors such as noisy and biased training data~\cite{liu2023mitigating, gunjal2023detecting, hu2023ciem}, limited visual perception~\cite{bai2023qwen, chen2023internvl}, misalignment between vision and language modalities~\cite{jiang2023hallucination}, and linguistic bias~\cite{guan2023hallusionbench, leng2023mitigating}. To mitigate these hallucination sources, various mitigation methods have been proposed. These include: creating clean and balanced datasets~\cite{liu2023mitigating, zhai2023halle}, scaling up visual resolution~\cite{liu2023improved, bai2023qwen} or utilizing additional perceptual models~\cite{yin2023woodpecker}, employing alignment modules with increased parameters~\cite{chen2023internvl, liu2023improved}, and designing decoding strategies that reduce over-reliance on language priors~\cite{leng2023mitigating}.

\begin{figure}[t]
  \centering
  \includegraphics[width=0.9\textwidth]{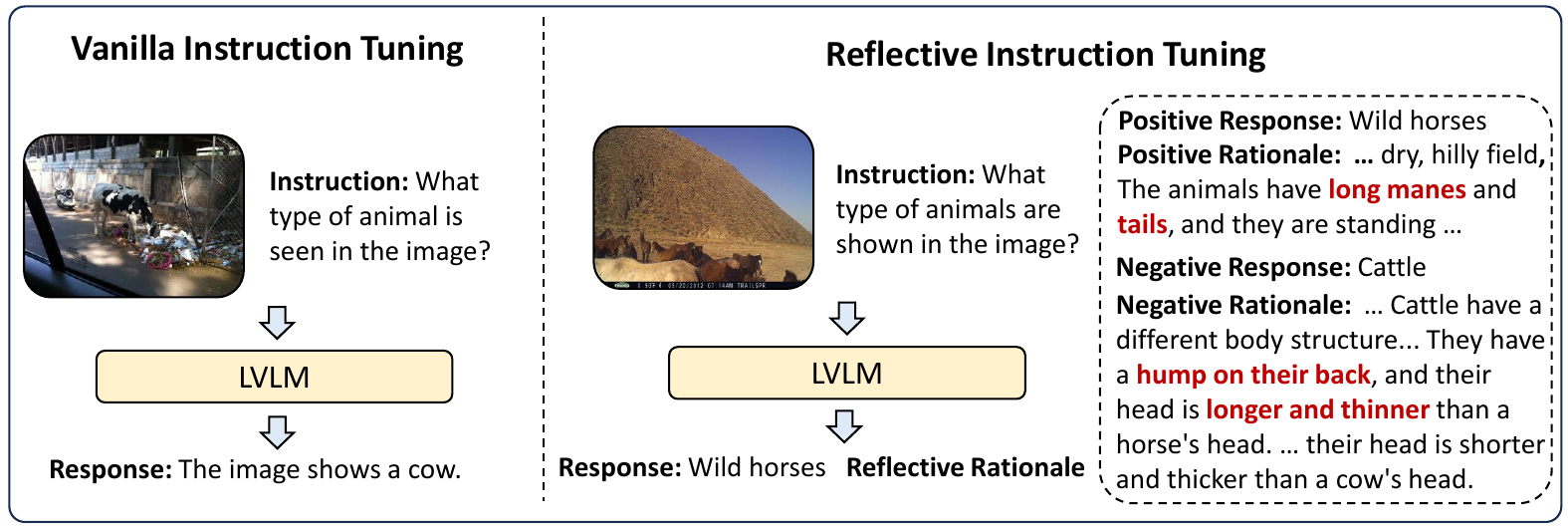}
  \caption{\textbf{Difference between vanilla instruction tuning and the proposed reflective instruction tuning.} Vanilla instruction tuning only trains LVLMs solely for response generation, lacking of supervising the learning of fine-grained reasoning details. Reflective instruction tuning additionally trains the model to reflect the rationale underlying the response, which provides more fine-grained supervision (\eg, the key visual evidence and facts to reach the response, highlighted in red), facilitating the model learning to capture more critical information.} 
  \label{fig:fig1}
\end{figure}

Despite significant progress, existing approaches often overlook a crucial factor contributing to hallucinations: the lack of fine-grained reasoning supervision. When learning to solve problems, humans often improve themselves by continuously reflecting on each step involved in the reasoning process. However, most previous LVLMs perform instruction tuning on datasets containing only instructions and direct responses, omitting access to the underlying rationale, \ie, intermediate reasoning steps. Specifically, Fig.~\ref{fig:fig1} (left) shows the training paradigm of vanilla instruction tuning applied to a recognition task. The model is trained solely to predict the animal category name, without explicit supervision on the reasoning details from the discriminating information (e.g., the defining attributes of the animal) to the judgment. This strategy risks training the model to establish superficial shortcuts between instructions and responses, rather than understanding the tasks and internalizing the underlying logic of the reasoning process. Consequently, the model becomes susceptible to producing incorrect inferences and hallucinatory content misaligned with visual content.

Inspired by the human ability to improve reasoning through reflection on intermediate steps~\cite{schunk1998self, boud2013reflection}, we propose reflective instruction tuning to address the aforementioned issue. Unlike vanilla instruction tuning, which only trains the model to predict the response, we incorporate the learning of underlying rationales, providing the model with fine-grained reasoning supervision. Specifically, reflective instruction tuning trains the model to generate positive rationales (justifying correct responses) and negative rationales (explaining errors in incorrect responses). The positive rationale explicitly provides more fine-grained visual content, the relationship between multiple visual concepts, and step-by-step reasoning. This facilitates the model to capture finer visual details and better understand the task definition underlying instruction. The negative rationale provides more illustration of the core characteristics linked to the erroneous categorization of negative answers, helping the model concentrate on crucial concepts within the decision-making process. This improves the model's ability to discriminate between correct and potential hallucinatory responses. Fig.~\ref{fig:fig1} (right) presents an example of positive and negative rationale, the positive rationale provides a more detailed justification of the correct answer, and the negative rationale highlights errors in the incorrect answer, aiding discrimination. Reflective instruction tuning mirrors the human cognitive process of reflection, in which individuals analyze step-by-step reasoning and errors to gain a deeper understanding. Training the model with both positive and negative rationales enhances several core cognitive capabilities: it improves visual perception through more visual information, facilitates instruction understanding with coherent reasoning logic, and promotes error awareness by highlighting critical information. Consequently, this training paradigm mitigates hallucinatory outputs.

While various visual instruction tuning datasets~\cite{liu2024visual, wang2023see, zhao2023svit, chen2023visual, liu2023mitigating, yin2024lamm} exist, most of them consist solely of instruction-response pairs, omitting the annotation of rationales. To facilitate reflective instruction tuning that learns from both positive and negative rationales, we introduce \textit{REVERIE}, the first large-scale visual instruction tuning dataset with ReflEctiVE ratIonalE annotations. \textit{REVERIE} builds upon existing VQA datasets~\cite{schwenk2022okvqa, lu2022learn}, and is further extended with natural images from Visual Genome~\cite{krishna2017visual}. The positive and negative rationales are annotated by Gemini family multimodal models~\cite{team2023gemini}. \textit{REVERIE} comprises 115k instructions encompassing four task types. Each instruction includes a positive response, an average of 1.2 negative responses, and accompanying rationales that justify the correctness or incorrectness of each response. This yields a total of 254k training instances, \ie, (instruction, response, rationale) tuples.

To enhance the adaptability of reflective instruction tuning to diverse tasks, we formulate the rationale data as a conversation format and position rationale learning subsequent to the model's prediction of responses, thereby decoupling rationale generation from the context of response prediction. We perform reflective instruction tuning on two LVLMs with the proposed \textit{REVERIE} dataset and further conduct experiments on multiple commonly used LVLM evaluation benchmarks. {The results show that the proposed reflective tuning with \textit{REVERIE} could yield substantial performance gain over the baselines. We also observed that utilizing the negative rationale could bring additional improvement than merely using the positive rationale.} In summary, our contributions are three-fold: (a) We propose reflective instruction tuning, a novel approach that incorporates learning from both positive and negative rationales into visual instruction tuning to mitigate hallucination. (b) We propose \textit{REVERIE}, the first large-scale instruction tuning dataset with both positive and negative fine-grained rationales. We hope this dataset could facilitate future research on rationale learning in LVLMs.  (c) Extensive experiments show that reflective instruction tuning with the proposed dataset could bring substantial performance gain across multiple LVLM evaluation benchmarks, demonstrating the effectiveness of incorporating reflective rationales for mitigating hallucinations.
\section{Related Work}

\noindent{\textbf{Large vision-language models.}}
The remarkable generalization capabilities of LLMs~\cite{touvron2023llama, zheng2023judging, touvron2023llama2, zhang2022opt, chowdhery2023palm, raffel2020exploring, chung2022scaling, ouyang2022training} have inspired the LVLMs~\cite{zhu2023minigpt, liu2024visual, zhang2023llavar, liu2023improved, instructblip, ye2023mplug, chen2023shikra, bai2023qwen}. LVLMs aim to extend the reasoning abilities of LLMs into the visual domain, potentially enabling more sophisticated and multimodal understanding. Typically, the architecture of previous LVLMs incorporates three fundamental components: a pre-trained visual encoder (\eg, CLIP~\cite{radford2021learning}) for visual perception, a pre-trained LLM (\eg, LLaMA~\cite{touvron2023llama}) serving as a language understanding and generation interface, and a connector (\eg, Q-former~\cite{li2023blip, instructblip}, linear projector~\cite{liu2024visual, zhu2023minigpt, liu2023improved}) facilitating vision-language alignment. A common training paradigm for these models involves vision-language pre-training using large-scale vision-text datasets~\cite{lin2014microsoft, schuhmann2022laion, zhu2023minigpt, liu2024visual} and visual instruction tuning~\cite{liu2024visual} with various visual task datasets~\cite{liu2024visual, liu2023improved, zhang2023llavar, han2023chartllama}. To enhance the generalization capabilities of models, previous methods have explored several techniques, such as: 1) employing stronger perception models~\cite{zong2024mova, jain2024vcoder} or LLMs~\cite{hanoona2024LLaVA++, internlmxcomposer}; 2) designing improved connectors~\cite{chen2023internvl, cha2024honeybee, lee2024meteor}; 3) utilizing a mixture-of-expert design~\cite{lin2024moe, chen2024llava}. Given the importance of visual instruction tuning in building general-purpose visual assistants, many studies have also focused on improving performance through the creation of high-quality instruction tuning datasets. These datasets target a wide range of visual understanding capabilities, encompassing text-rich image comprehension~\cite{zhang2023llavar}, region-level analysis~\cite{zhao2023chatspot}, chart interpretation~\cite{han2023chartllama}, and multi-image reasoning~\cite{huang2023sparkles}. The proposed \textit{REVIRIE} dataset also aims to improve the model performance during visual instruction tuning, but we focus on enhancing the reasoning capability by offering fine-grained reasoning supervision.

\noindent{\textbf{Learning from reasoning steps.}} 
Several studies~\cite{wang2022self, wei2022chain, ge2023chain, lu2022learn, wang2022rationale, lightman2023let} highlight the effectiveness of learning from detailed reasoning steps. Wei et al.~\cite{wei2022chain} propose chain-of-thought prompting, where LLMs are explicitly prompted to think step-by-step during inference, leading to substantial gains on reasoning tasks. Furthermore, recent work suggests that training with explicit rationales can improve model reasoning capabilities~\cite{zelikman2022star, magister2022teaching}. Additionally, learning from answer explanations has shown promise in enhancing performance on vision-language reasoning tasks ~\cite{lu2022learn, wu2023role, zhang2023multimodal}. Despite these advancements, learning from reasoning steps in visual instruction tuning remains under-explored due to the scarcity of large-scale rationale-annotated visual instruction tuning datasets.  LLaVA-Instruct-158k~\cite{liu2024visual} contains a portion of annotations for complex reasoning tasks, and the responses have step-by-step reasoning process annotation that is similar to the concept of rationale. However, certain limitations remain: 1) these rationales are all positive coupled with correct responses, neglecting negative rationales crucial for discerning incorrect responses, and 2) they are restricted to open-ended reasoning tasks, offering limited benefits to the diverse range of tasks. We address these gaps by proposing \textit{REVERIE}, the first large-scale visual instruction tuning dataset that contains both positive and negative rationales on various visual tasks, facilitating the future research on rationale learning in visual instruction tuning.

\noindent{\textbf{Hallucination in LVLMs.}} Previous research on hallucination in LVLMs has focused on two key areas: 1) the development of benchmarks for evaluating hallucinations~\cite{guan2023hallusionbench, li2023evaluating, sun2023aligning, liu2023mitigating}, and 2) the exploration of techniques to mitigate hallucinations~\cite{liu2023mitigating, wang2024mitigating, leng2024mitigating, jiang2023hallucination}. Li et al.~\cite{li2023evaluating} identify the object hallucination issue in LVLMs, and they propose the POPE benchmark for its evaluation. Guan et al.~\cite{guan2023hallusionbench} divide the hallucination into language hallucination and visual hallucination and further introduce HALLUSIONBENCH to evaluate and diagnose each hallucination. Various studies address hallucination mitigation. Liu et al.~\cite{liu2023mitigating} reduce hallucinations with a balanced visual instruction tuning dataset to better handle negative instructions. Jiang et al.~\cite{jiang2023hallucination} employ contrastive learning to distinguish hallucinated and non-hallucinated representations in latent space. Wang et al.~\cite{wang2024mitigating} and Leng et al.~\cite{leng2024mitigating} design contrastive decoding strategies in inference to mitigate the hallucination caused by language bias. Similar to~\cite{liu2023mitigating}, our work also focuses on mitigating hallucinations through improved visual instruction datasets. However, we emphasize fine-grained reasoning supervision to enhance consistency throughout the reasoning process.

\section{The REVERIE Dataset}
\begin{figure}
  \centering
  \includegraphics[width=0.9\textwidth]{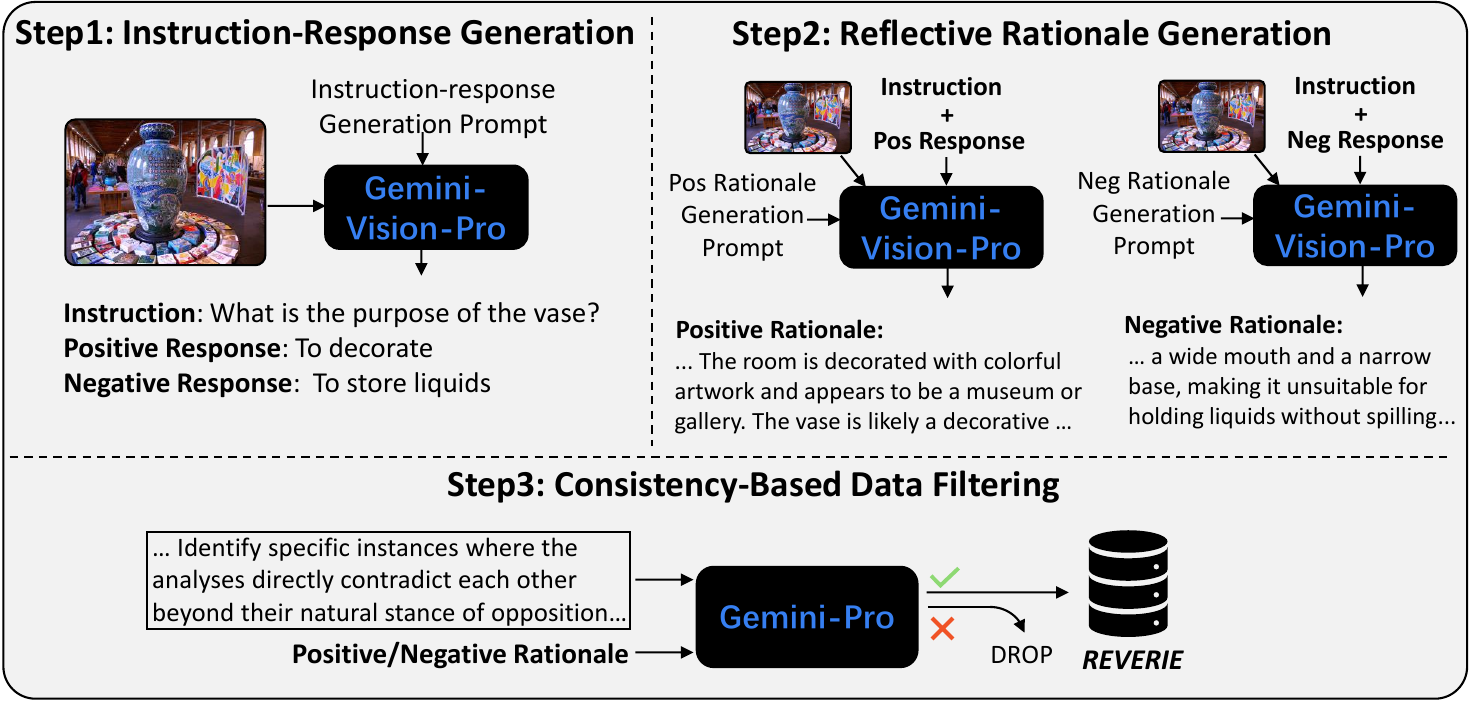}
  \caption{\textbf{Overview of the REVERIE dataset's data collection pipeline.} We first employ Gemini-Vision-Pro to annotate the instructions, responses, and rationales for each image. Gemini-Pro is then used to check the consistency between positive and negative rationales. Inconsistent samples are filtered to maintain dataset quality.} 
  \label{fig:fig2}
\end{figure}

Despite previous works have provided many diverse visual instruction tuning datasets (\eg, LLaVA-Instruct-158k~\cite{liu2024visual}, LRV-Instruction~\cite{liu2023mitigating}), only a small fraction of samples possess rationale annotations for positive responses. Furthermore, these datasets often omit annotations for hard negative responses and their underlying rationales, rendering them hard to provide fine-grained and discriminative supervision for addressing hallucinations. To facilitate model ``reflect'' from both positive and negative responses, we propose \textit{REVERIE}, the first large-scale visual instruction tuning dataset with detailed reflective rationale annotation, serving as a basis of the proposed reflective instruction tuning strategy (introduced in Sec.~\ref{sec: reflective}). As shown in Table~\ref{tab:dataset_compare}, \textit{REVERIE} provides hard negative response annotation and contains more extensive rationale annotations compared with the existing datasets. Fig.~\ref{fig:fig2} shows the overall pipeline of our data collection strategy, we first annotate instructions, responses, and positive/negative rationales for each collected image, followed by filtering out noisy samples by the consistency-based filtering strategy. The details are as follows (The prompts for data generation are available in supplementary materials).

\begin{table}
\centering
\caption{\textbf{Comparison of the \textit{REVERIE} with the data used by other LVLMs.}}
\scalebox{0.87}{
\begin{tabular}{lccccc}
\toprule
& REVERIE & LLaVA-1.5 & InstructBLIP & MiniGPT-4 & LRV-Instruction \\ 
\midrule
\# Training Instances & 254k & 665k & 1.2M & 3.5k & 400k \\
Hard Negative Responses? & \textcolor{green}{\ding{52}} & \textcolor{red}{\ding{56}} & \textcolor{red}{\ding{56}} & \textcolor{red}{\ding{56}} & \textcolor{red}{\ding{56}} \\
\# Positive Rationales & 115k & 77k & 77k & \textcolor{red}{\ding{56}} & \textcolor{red}{\ding{56}} \\
\# Negative Rationales & 138k & \textcolor{red}{\ding{56}} & \textcolor{red}{\ding{56}} & \textcolor{red}{\ding{56}} & \textcolor{red}{\ding{56}} \\
\bottomrule
\end{tabular}
}
\label{tab:dataset_compare}
\end{table}
\noindent{\textbf{Data collection and annotation.}} To ensure the dataset encompasses diverse visual concepts and broad domain coverage, we sample a subset of images from the Visual Genome dataset~\cite{krishna2017visual}. Subsequently, we annotate these images with instructions and corresponding responses. Given the high cost of manual labelling, we leverage a strong LVLM, Gemini-Vision-Pro, as our annotator. Considering that the rationale for addressing simple query instruction may contain limited intermediate steps and benefit less to learning fine-grained reasoning, we explicitly prompt Gemini to generate instructions that need multi-step to solve and avoid generating simple queries. Additionally, it is prompted to generate both positive and negative responses in conjunction with each instruction. Intuitively, a more confusing negative response demands more discriminative information to distinguish it from the positive answer. Thus, we explicitly prompt Gemini to generate negative responses that are easy to confuse. 

After generating the instructions and responses, we prompt Gemini-Vision-Pro to generate rationales underlying the responses. Considering that positive and negative rationales have different emphases, we design two prompts and generate them separately. The positive rationales are expected to help the model learn the inherent knowledge and visual concepts underlying each problem-solving step, thereby facilitating fine-grained vision-language alignment and understanding of task definitions. It should include 1) analysis of visual content and instructions, 2) identification of core visual concepts relevant to the instructions, and 3) step-by-step reasoning grounded on the visual information and required knowledge. This is achieved by explicitly incorporating words such as ``explanation'', ``details'', and ``knowledge'' into the prompt to facilitate Gemini-Vision-Pro to include key image details and necessary outside knowledge in the rationale. The negative rationales are expected to provide discriminative information that aids in distinguishing between valid and hallucinatory outputs. Thus, we let Gemini-Vision-Pro highlight discriminative details in the negative rationale.

Consider A-OK-VQA~\cite{schwenk2022okvqa} and ScienceQA~\cite{lu2022learn} already have some human annotated QA pairs and positive rationales, we include them into the \textit{REVERIE} dataset to increase task diversity. Since both datasets provide questions in a multiple-choice format, we incorporate all incorrect answer choices and generate an additional negative rationale for each of them with the same pipeline.

\noindent{\textbf{Consistency-based data filtering.}}
While Gemini-Vision-Pro demonstrates impressive capabilities, it can still generate hallucinatory content when tasked to generate responses and rationales. Previous work~\cite{yu2023hallucidoctor} has shown that noise training data is one of the sources of the hallucinations. To guarantee the quality of generated data, we propose a method that leverages the consistency between positive and negative rationales to filter out noisy examples. Since both positive and negative rationales explicate the reasoning process behind the instruction but emphasize different aspects, they should contain information that is mutually consistent. When one rationale includes noisy information, this inherent consistency will be disrupted. Inspired by that, we employ a large language model, \ie, Gemini-Pro, to detect inconsistencies between these two types of rationales and only keep the samples that contain intra-consistent information (See supplementary materials for detailed prompt).

\begin{figure}
\centering
  \includegraphics[width=0.9\textwidth]{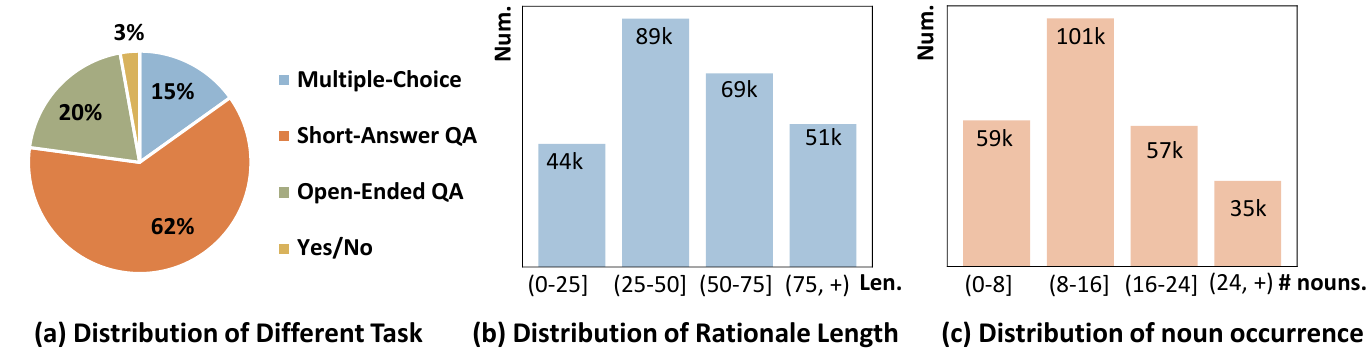}
  \caption{\textbf{Statistics of the \textit{REVERIE} dataset.}} 
  \label{fig:statistics}
\end{figure}

\noindent{\textbf{Data statistics.}} After the removal of noisy samples, the \textit{REVERIE} dataset comprises 71,558 natural images. This includes 50,938 images sourced from Visual Genome~\cite{krishna2017visual}, 15,706 from the COCO ~\cite{lin2014microsoft} and 4,914 images from ScienceQA~\cite{lu2022learn}. \textit{REVERIE} contains 115,280 instructions paired with corresponding positive responses and 138,897 negative responses, where each response is supplemented with a reflective rationale, rendering a total of 254,177 training instances. According to the output formats, \textit{REVERIE} covers four types of vision-language tasks, including multiple-choice QA, short-answer QA, open-ended QA, and Yes/No questions, and their distributions are visualized in Fig.~\ref{fig:statistics} (a). We also analyzed the distribution of rationale length and the frequency of noun occurrence. As shown in Fig.~\ref{fig:statistics} (b) and (c), over 50\% of rationales exceed 25 words in length and contain more than 8 nouns, suggesting that rationales provide a substantial amount of information and knowledge.

\section{Reflective Instruction Tuning}
\label{sec: reflective}

Reflection plays a crucial role in the human learning process, allowing individuals to extract meaningful insights, learn from their mistakes, and make informed choices in the future. However, vanilla instruction tuning solely focuses on training models to predict responses, thus omitting access to the underlying rationale. This limitation impedes the model's ability to effectively internalize and logically process instruction-relevant information, increasing the likelihood of generating hallucinatory outputs. Inspired by humans' ability to improve problem-solving skills through reflection on explanations for both correct and incorrect answers~\cite{schunk1998self, boud2013reflection}, we propose reflective instruction tuning. This approach facilitates learning from the rationale behind responses, extending the training process to predict both positive and negative rationales rather than merely generating correct responses to instructions.

\noindent\textbf{Overall architecture.} The objective of visual instruction tuning is to improve the instruction-following capability of the LVLM by leveraging diverse visual task data. Typically, LVLMs consist of a visual encoder, a vision-language connector, and an LLM. Given an image sample $I$ and user instruction $X$, the visual encoder extracts image token features, which are then projected into the language space by the modality connector. The projected visual tokens, along with the textual instructions, serve as the prefix in the LLM. An auto-regressive LLM is trained to generate response $A$ that aligns with both the instructions and the image.
 
\noindent\textbf{Multi-turn conversation.}
Directly training a model to concurrently predict rationales alongside responses in a single-turn prediction might hinder inference results. This is because some tasks do not necessitate explicit rationale generation, limiting its real-world applicability. To address this, we decouple rationale learning from response prediction to improve the adaptability of reflective instruction tuning to diverse tasks. We achieve this by formulating the rationale data in a multi-turn conversational format. The first turn incorporates $I$ and $X$ as inputs, prompting the model to generate $A$. Subsequently, the second turn introduces rationale generation prompts to let the model predict rationale. By putting the rationale learning behind, response prediction relies solely on the context of $I$ and $X$, mitigating interference to the inference results. To make the model understand this rationale generation task, we explicitly prompt the model to generate knowledge and facts that are related to instruction. Specifically, in positive rationale learning, the model receives the prompt ``\textvtt{Explain why. Including any necessary facts or knowledge}'' following its response. In negative rationale learning, an incorrect answer is provided, followed by the prompt ``\textvtt{Explain why this answer is wrong: \{incorrect answer\}. Including any necessary facts or knowledge.}''
\section{Experiments}
We demonstrate the effectiveness of learning from rationales by applying reflective instruction tuning with the \textit{REVERIE} dataset to two baseline LVLMs:  (1) LLaVA-1.0-7b-lora~\cite{liu2024visual}: We initialize the model pre-trained with LAION-CC-SBU-558k, and perform instruction tuning with the combination of \textit{REVERIE} and LLaVA-Instruct-80k. (2) LLaVA-1.5-7b-lora~\cite{liu2023improved}: Similar to LLaVA-1.0, we perform instruction tuning with the combination of \textit{REVERIE} and the original 665k instructions. The resulting models are denoted as REVERIE-1.0-7b-lora and REVERIE-1.5-7b-lora, respectively. Implementation details are provided in the supplementary materials.



\subsection{Main Results}

\noindent{\textbf{Performance comparison with baselines on six benchmarks.}} We compare the performance of the model reflective tuned with the proposed \textit{REVERIE} dataset. As shown in Table~\ref{tab:results}, training with \textit{REVERIE} yields clear performance gain on most of the benchmarks across 2 baselines. Notably, we achieve a score of 83.8 on the POPE benchmark, reflecting a significant improvement of 12.7 points over the baseline LLaVA-1.0-7b-lora. This suggests that the \textit{REVERIE} dataset provides more fine-grained supervision, enabling the model to make more accurate judgments regarding object existence and significantly alleviating hallucination issues. Additionally, we observe substantial performance gains on two comprehensive LVLM evaluation benchmarks, MME and MMBench. Specifically, REVERIE-1.0-7b-lora achieves improvements of 321.2 and 27.2 points on MME and MMBench, respectively, compared to the baseline. This suggests that 1) the \textit{REVERIE} dataset encompasses a wider range of visual concepts and knowledge compared to the original LLaVA-Instruct-80k dataset, demonstrably benefiting performance on diverse visual perception tasks evaluated in MME; 2) incorporating rationales during training empowers the model to develop a deeper understanding of task definitions and effectively distinguish positive answers from provided negative answers, thereby achieving superior performance on multiple-choice question benchmarks like MMBench. Notably, for a fair comparison, we trained the baseline, LLaVA-1.5-7b-lora, using the combination of its original instruction tuning data and the VQA data from ScienceQA (only data with image context). We observed that simply introducing the ScienceQA QA data led to a performance decline on MME and GQA. This suggests that learning from response data alone may establish unintended shortcuts between instructions and responses, particularly for the complex reasoning instructions in ScienceQA. Our REVERIE-1.5-7b-lora demonstrates significant performance gains over its baseline on ScienceQA, MME, and GQA. Meanwhile, it achieves competitive performance on other benchmarks. This suggests that the rationale from our dataset may help alleviate the shortcut issue.

\begin{table}[t]
\caption{\textbf{Comparison with baselines on six benchmarks.} $^{*}$ means training images of the dataset are used during training. $^{\dagger}$ means rerun using their released code.}
\centering
\scalebox{0.87}{
\begin{tabular}{l | cccccc }
\toprule
Method & ScienceQA$^\text{I}$ & POPE & MME & MMBench & MM-Vet & GQA \\
\midrule
InstructBLIP (vicuna-7b)  & 60.5 & -- & -- & 36 & 26.2 & 49.2\\
Qwen-VL-7b & 67.1 & -- & -- & 38.2 & -- & 59.3 \\
Qwen-VL-Chat-7b & 68.2 & -- & 1487.5 & 60.6 & -- & ~57.5$^{*}$ \\
LLaVA-1.5-7b & 66.8 & 85.9 & 1510.7 & 64.3 & 30.5 & ~62.0$^{*}$ \\
LLaVA-1.5-7b-lora & 68.4 & 86.4 & 1476.9 & 66.1 & 30.2 & ~63.0$^{*}$ \\
LLaVA-1.5-7b-lora$^{\dagger}$ & 67.7 & 86.1 & 1471.0 & 65.1 & 32.5 & ~62.8$^{*}$ \\
\midrule
LLaVA-1.0-7b-lora$^{\dagger}$ & 42.7 & 71.1 & 819.8 & 27.2 & 30.0 & 7.10 \\
REVERIE-1.0-7b-lora & ~70.1$^{*}$ & 83.8 & 1168.1 & 55.4 & 27.8 & 36.5 \\
\midrule
LLaVA-1.5-7b-lora (w/ ScienceQA)$^{\dagger}$ & ~76.3$^{*}$ & 86.6 & 1439.2 & 67.9 & 31.1 & ~60.7$^{*}$ \\
REVERIE-1.5-7b-lora & ~80.5$^{*}$ & 86.4 & 1474.9 & 67.3 & 30.8 & ~61.8$^{*}$ \\

\bottomrule
\end{tabular}
}
\label{tab:results}
\end{table}

\begin{table*}[t] 
\caption{\textbf{Results of using different rationales on POPE.}}
\centering
\begin{tabular}{p{20mm} p{35mm} ccc |c}
\toprule
 \textbf{Subset} &\textbf{Model} & Accuracy & Precision & Recall & F1 Score\\
\midrule
\multirow{4}{*}{\textit{Random}} & Response only  & 86.87   & 89.42   & 84.53  & 86.90   \\ 
    & w/ pos rationales  & 85.60   & 87.93   & 83.53   & 85.68   \\
    & w/ neg rationales     & 86.25   & 87.98   & \textbf{84.93}   & 86.43  \\
    & w/ pos + neg  rationales  & \textbf{87.15}   & \textbf{90.21}   & 84.20   & \textbf{87.10}  \\
\midrule
\multirow{4}{*}{\textit{Popular}} & Response only & 82.57   & 81.33   & 84.53   & 82.90  \\ 
   & w/ pos rationales    & 81.80   & 80.73   & 83.53   & 82.11 \\
  & w/ neg  rationales    & 83.30   & 82.25   & \textbf{84.93}   & 83.57  \\
   & w/ pos + neg rationales   & \textbf{84.00}   & \textbf{83.86}   & 84.20   & \textbf{84.03} \\
\midrule
\multirow{4}{*}{\textit{Adversarial}} & Response only & 76.83   & 73.25   & 84.53   & 78.49\\ 
  & w/ pos rationales & 77.23   & 74.19   & 83.53   & 78.58\\
   &  w/ neg rationales & 77.73   & 74.24  & \textbf{84.93}   & 79.23\\
   & w/ pos + neg rationales & \textbf{79.37}   & \textbf{76.78}   & 84.20   & \textbf{80.31}  \\

\bottomrule
\end{tabular}
\label{tab:POPE}
\end{table*}

\noindent{\textbf{Effectiveness of positive/negative rationale.}} We conduct experiments to analyze the role of different rationales used in reflective instruction tuning. Table~\ref{tab:POPE}, Table~\ref{tab:MMB} and Table~\ref{tab:GQA} show the results of using different rationales on POPE, MMBench, and GQA, respectively. The results show that using both positive and negative rationale yields the best performance on most metrics. The results on POPE show that using both two rationales brings the highest F1 score, suggesting that training with rationale may alleviate the instruction-response shortcut and further alleviate the object hallucination issue. The results on MMBench show that training exclusively with positive rationales yields higher perception scores (in terms of CP and FP-C). This indicates that positive rationales enhance the capture of visual details. Conversely, training solely with negative rationales results in superior reasoning performance (in terms of AR and RR), suggesting that negative rationales provide critical clues that improve discriminative ability. We also infer that training with negative rationales could help the model better distinguish the correct answer from the negative response, benefiting more to the task with candidate answers provided, \eg, multiple choice. Our results on GQA further show that both positive and negative rationales contribute to performance gains in questions with category as their subject. This suggests that rationales provide more fine-grained information that enables the model to better understand the defining characteristics of a category. 

\begin{table*}[t] 
\caption{\textbf{Results of using different rationales on MMBench.} We report L2-ability performance, where AR, CP, FP-S, FP-C, LR, and RR denote attribute reasoning, coarse perception, fine-grained perception (single-instance), fine-grained perception (cross-instance), logic reasoning, and relation reasoning, respectively.}
\centering
\begin{tabular}{p{33mm} p{9mm} p{9mm} p{11mm} p{11mm} p{9mm} p{9mm}| c}
\toprule
  \textbf{Model} & AR & CP & FP-S & FP-C & LR & RR & Accuracy\\
\midrule
  Response only  & 61.69  & 60.09 & 50.51 & 42.76 & \textbf{26.67} & 42.61 & 51.70   \\ 
    w/ pos  rationales    & 63.18   & 67.11   & 44.14   & 49.16 & 24.17 & 40.87 & 52.13   \\
    w/ neg rationales  & 64.68   & 64.43   & 50.84   & 44.83 & 22.50 & 43.48 & 52.30   \\
    w/ pos + neg rationales & \textbf{65.14} & \textbf{68.46} & \textbf{57.24} & \textbf{51.52} & 19.17 & \textbf{50.43} & \textbf{55.44} \\
\bottomrule
\end{tabular}
\label{tab:MMB}
\end{table*}

\begin{table*}[t] 
\centering
\caption{\textbf{Results of using different rationales on GQA.} We report the semantic-level accuracy to analyze the impact of rationale on questions with different subjects.}
\begin{tabular}{lccccc|c}
\toprule
  \textbf{Model} & Attribute & Object & Relation & Global & Category & Accuracy\\
\midrule
  Response only  & 47.17   & \textbf{59.90}   & 22.21   & \underline{31.21} & 19.67 & 34.71   \\ 
    w/ pos rationales  & 46.78   & 57.46   & 23.44   & 29.30 & 22.89 & 35.19   \\
    w/ neg rationales  & \textbf{48.01}   & 56.68   & \textbf{24.77}   & \textbf{31.85} & \textbf{28.63} & \textbf{36.77}   \\
    w/ pos + neg rationales & \underline{47.32} & \underline{58.35} & \underline{24.76} & \textbf{31.85} & \underline{28.11} & \underline{36.53} \\
\bottomrule
\end{tabular}
\label{tab:GQA}
\end{table*}

\subsection{Ablation Studies}

We investigate several key designs of reflective instruction tuning and \textit{REVERIE}: 1) the design of rationale generation prompt; 2) the effectiveness of consistency-based data filtering; 3) the design of conversational context; All experiments are conducted with REVERIE-1.0-7b-lora.

\noindent{\textbf{Rationale generation prompt.}} We conducted ablation studies to analyze the design of the rationale generation prompt employed in reflective instruction tuning. Specifically, we test the variants of rationale generation prompt from three perspectives: (1) the use of explicit guidance, \eg, ``\textvtt{Including any necessary facts or knowledge}''; 2) the utilization of separate prompts for positive and negative rationales; 3) the use of length-specific prompts. Table~\ref{tab:prompt} summarizes the performance of these prompt variants. Our final choice, represented by Line (d), includes only explicit guidance and yields the best overall performance across four benchmarks. Line (a) demonstrates that using the prompt without explicit guidance leads to a significant performance drop on the MME benchmark. This suggests that providing clear guidance for clarifying the rationale generation task is crucial for aligning rationales with visual content. Moreover, Lines (c) and (d) indicate that overly complex prompts containing excessive constraints may impede the model's understanding of the rationale generation task, adversely affecting performance across most benchmarks.

\begin{table}[t] 
\caption{\textbf{Performance comparison of different rationale prompts.}}
\centering
\begin{tabular}{l|c|c|c|cccc}
\toprule
 & Explicit Guidance & Type-Specific & Length Control & MMB & POPE & GQA & MME \\
\midrule
(a) &  &  &  & 55.4 & 83.8 & \textbf{36.8} & 1128.6 \\
(b) & \checkmark & \checkmark & & 55.2 & 83.2 & 36.0 & 1111.9 \\
(c) & \checkmark & \checkmark & \checkmark & 54.3 & 83.8 & 35.8 & 1141.0 \\
(d) & \checkmark &  & & \textbf{55.4} & \textbf{83.8} & 36.5 & \textbf{1168.1} \\
\bottomrule
\end{tabular}
\label{tab:prompt}
\end{table}

\noindent{\textbf{Consistency-based data filtering.}} We conducted experiments on the noise-infused rationale data to investigate the effectiveness of the proposed consistency-based data filtering strategy. Specifically, we train the model with the rationale data before filtering. As shown in Table~\ref{tab:noise}, training with noised data that are not intra-consistent resulted in performance degradation across most benchmarks, suggesting that using the consistency-based filter strategy to maintain data quality is necessary for learning from rationale.

\begin{table}
  \centering
    \scriptsize
~ 
\begin{minipage}{0.48\linewidth}
\caption{\textbf{Effect of rationale data filtering on model performance.}}
\centering
\begin{tabular}{l|cccc}
\toprule
 \textbf{Model} & MMB & POPE & GQA & MME \\
\midrule
w/o filtering & 55.2 & 83.5 & 36.3 & 1132.3 \\
w/ filtering & \textbf{55.4} & \textbf{83.8} & \textbf{36.5} & \textbf{1168.1} \\
\bottomrule
\end{tabular}
\label{tab:noise}
\end{minipage}
~
\begin{minipage}{0.48\linewidth}
\caption{\textbf{Performance comparison of different conversational contexts.}}
\centering
\begin{tabular}{l|cccc}
\toprule
 \textbf{Model} & MMB & POPE & GQA & MME \\
\midrule
pos rationale first & 49.3 & 79.5 & 36.2 & 1144.7 \\
neg rationale first & 48.6 & 82.4 & 36.4 & 1160.2 \\
separate context & \textbf{55.4} & \textbf{83.8} & \textbf{36.5} & \textbf{1168.1} \\
\bottomrule
\end{tabular}
\label{tab:context}
\end{minipage}
\end{table}

\noindent{\textbf{Conversational context of reflective instruction tuning.}} We conducted ablation studies on several training data format variants, each incorporating different conversational contexts. We test the performance of three variants: 1) positive rationale first, where all negative rationales are learned to predict after positive rationale turns; 2) negative rationale first, the inverse order of the previous one; 3) separate context, where the positive and negative rationales are separate into two distinct samples. The previous two variants put the positive and negative rationales for one instruction into a single conversation, making the model see previous rationales when learning to predict the current one. Table~\ref{tab:context} shows the performance of these three variants. Our final choice that using separate contexts consistently outperforms the other two variants across four benchmarks. This suggests that providing the model with access to additional rationales during the current rationale generation step might lead to information leakage. In such a scenario, the model may tend to leverage pre-existing information from other rationales rather than learning to reason according to visual information and acquired knowledge. This hinders the model benefited from the fine-grained supervision from rationales.

\subsection{Qualitative Analysis}

\begin{figure}
\centering
  \includegraphics[width=0.85\textwidth]{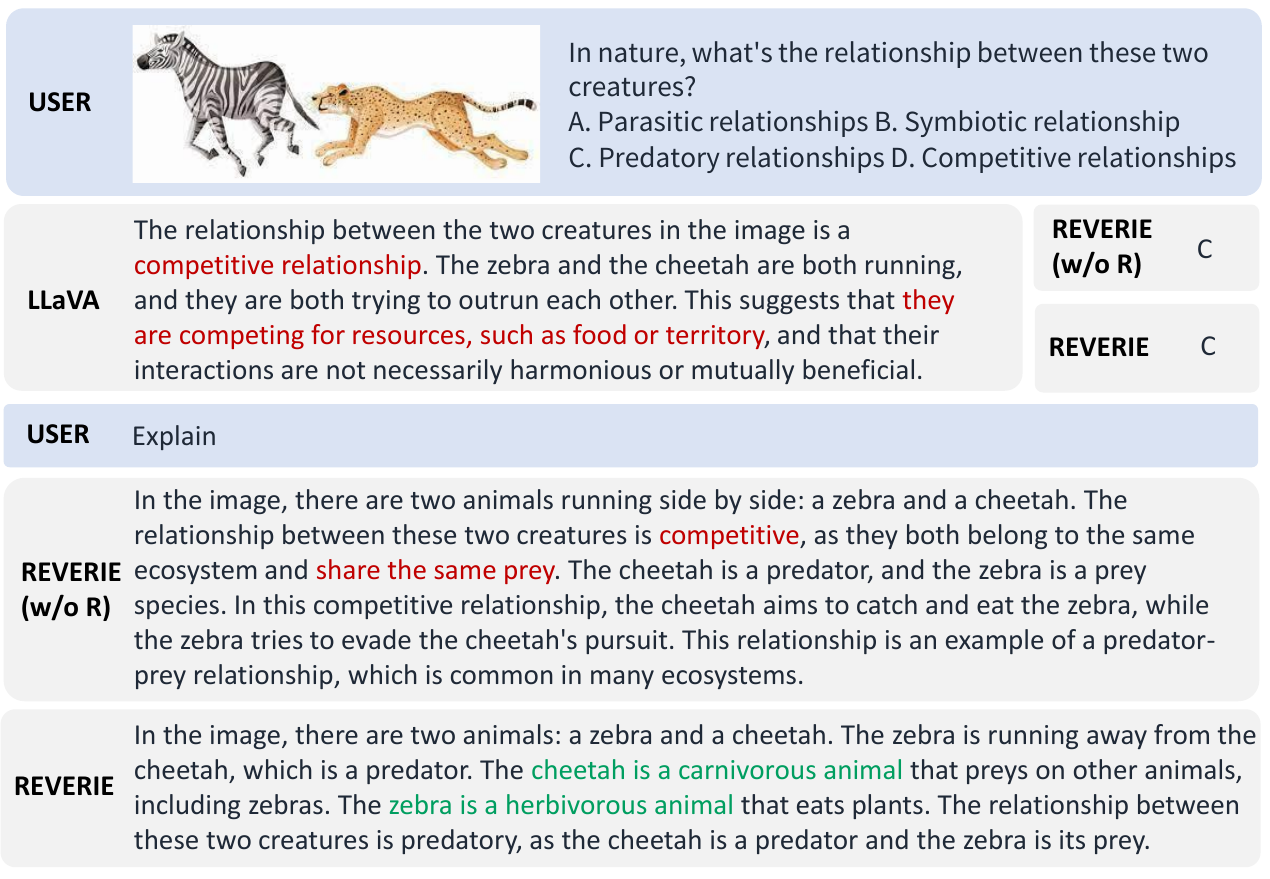}
  \caption{\textbf{Visualization of the generation of positive rationales.}} 
  \label{fig:pos_examples}
\end{figure}

\begin{figure}
\centering
  \includegraphics[width=0.85\textwidth]{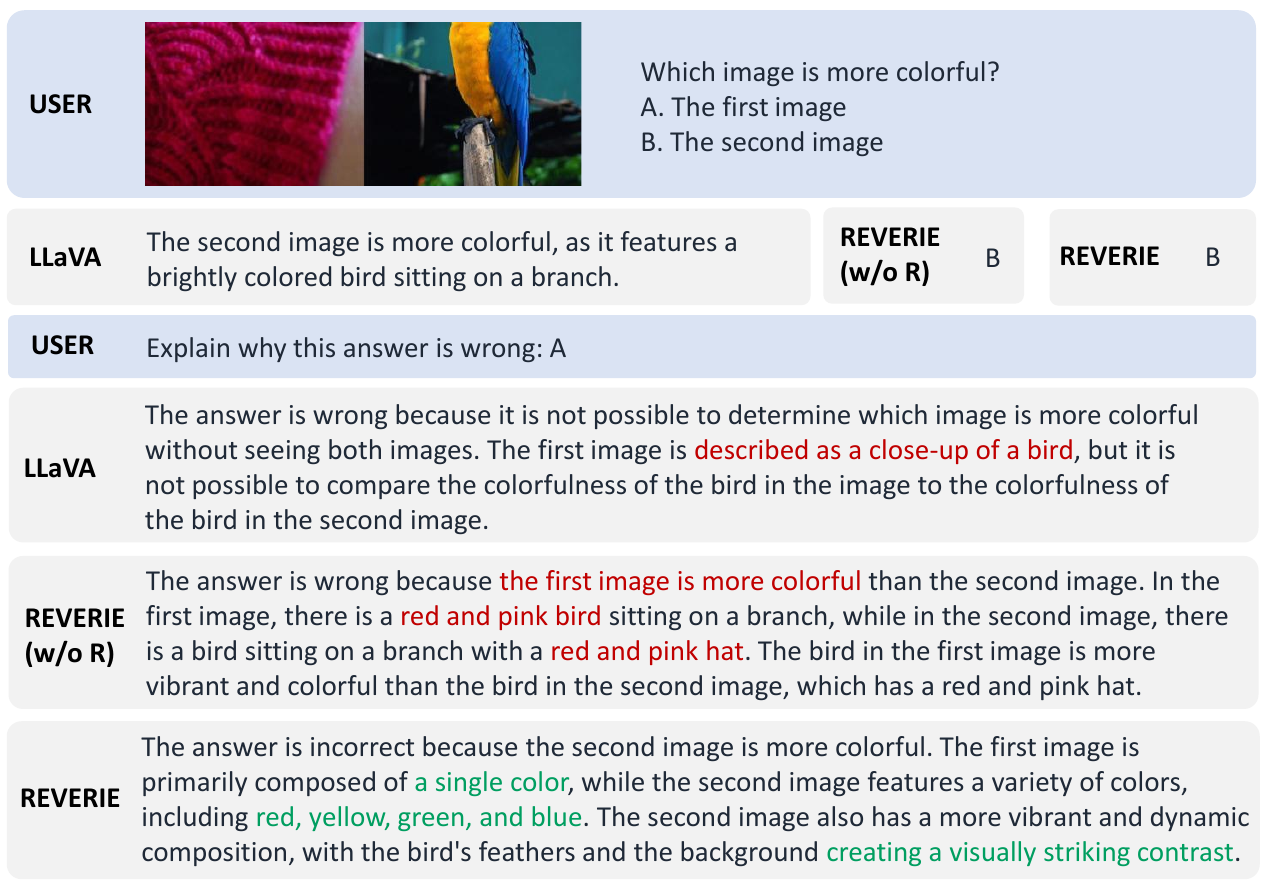}
  \caption{\textbf{Visualization of the generation of negative rationales.}} 
  \label{fig:neg_examples}
\end{figure}

We conducted a case analysis to assess whether reflective instruction tuning improves fine-grained reasoning and mitigates hallucination. We compared baseline models with our proposed approach by prompting both to generate explicit rationales for their responses, aiming to verify if the models' logic aligns with correct reasoning. Fig.~\ref{fig:pos_examples} visualizes the output rationales of different models. LLaVA directly generates rationale along with the response. However, the response is incorrect, and the rationale contains hallucinated content (highlighted in red). REVERIE (w/o R) (\ie, the model trained without rationales) and REVERIE both predicted the correct answer. However, REVERIE (w/o R) generates a rationale inconsistent with the response, suggesting that training without fine-grained supervision may lead to the shortcut between instruction and response rather than solving the problem from the correct reasoning paths. In contrast, the rationale generated by REVERIE presents more information that is necessary (highlighted in green) for addressing the instruction. We further evaluate the models' capability to generate negative rationales. As shown in Fig.~\ref{fig:neg_examples}, REVERIE correctly identifies crucial information (\eg, all presented colors) from images, while the other two models are misled by the context of two sub-images. This suggests that incorporating rationale as supervision enhances the model's fine-grained understanding of visual context, reducing hallucination.
\section{Conclusion}
In this work, we identify a key contributor to hallucinations in existing LVLMs: the lack of fine-grained reasoning supervision during visual instruction tuning. To address this, we introduce reflective instruction tuning, which draws inspiration from human reflective learning processes by training the model on the rationales of positive and negative answers. Additionally, we propose \textit{REVERIE}, the first large-scale visual instruction tuning dataset featuring comprehensive annotations for both positive and negative rationales. Experiments demonstrate that reflective instruction tuning with \textit{REVERIE} significantly enhances LVLM performance, emphasizing the importance of fine-grained reasoning supervision in mitigating hallucinations.
\section*{Acknowledgement}
This work is supported by the National Natural Science Foundation of China (Grant NO.62122035).

\bibliographystyle{splncs04}
\bibliography{ref}

\begin{thebibliography}{10}
\providecommand{\url}[1]{\texttt{#1}}
\providecommand{\urlprefix}{URL }
\providecommand{\doi}[1]{https://doi.org/#1}

\bibitem{bai2023qwen}
Bai, J., Bai, S., Yang, S., Wang, S., Tan, S., Wang, P., Lin, J., Zhou, C., Zhou, J.: Qwen-vl: A frontier large vision-language model with versatile abilities. arXiv preprint arXiv:2308.12966  (2023)

\bibitem{boud2013reflection}
Boud, D., Keogh, R., Walker, D.: Reflection: Turning experience into learning. Routledge (2013)

\bibitem{cha2024honeybee}
Cha, J., Kang, W., Mun, J., Roh, B.: Honeybee: Locality-enhanced projector for multimodal llm. In: Proceedings of the IEEE/CVF Conference on Computer Vision and Pattern Recognition. pp. 13817--13827 (2024)

\bibitem{chen2023visual}
Chen, D., Liu, J., Dai, W., Wang, B.: Visual instruction tuning with polite flamingo. arXiv preprint arXiv:2307.01003  (2023)

\bibitem{chen2023shikra}
Chen, K., Zhang, Z., Zeng, W., Zhang, R., Zhu, F., Zhao, R.: Shikra: Unleashing multimodal llm's referential dialogue magic. arXiv preprint arXiv:2306.15195  (2023)

\bibitem{chen2024llava}
Chen, S., Jie, Z., Ma, L.: Llava-mole: Sparse mixture of lora experts for mitigating data conflicts in instruction finetuning mllms. arXiv preprint arXiv:2401.16160  (2024)

\bibitem{chen2023internvl}
Chen, Z., Wu, J., Wang, W., Su, W., Chen, G., Xing, S., Muyan, Z., Zhang, Q., Zhu, X., Lu, L., et~al.: Internvl: Scaling up vision foundation models and aligning for generic visual-linguistic tasks. arXiv preprint arXiv:2312.14238  (2023)

\bibitem{chowdhery2023palm}
Chowdhery, A., Narang, S., Devlin, J., Bosma, M., Mishra, G., Roberts, A., Barham, P., Chung, H.W., Sutton, C., Gehrmann, S., et~al.: Palm: Scaling language modeling with pathways. Journal of Machine Learning Research  \textbf{24}(240),  1--113 (2023)

\bibitem{chung2022scaling}
Chung, H.W., Hou, L., Longpre, S., Zoph, B., Tay, Y., Fedus, W., Li, Y., Wang, X., Dehghani, M., Brahma, S., et~al.: Scaling instruction-finetuned language models. arXiv preprint arXiv:2210.11416  (2022)

\bibitem{instructblip}
Dai, W., Li, J., Li, D., Tiong, A.M.H., Zhao, J., Wang, W., Li, B., Fung, P., Hoi, S.: Instructblip: Towards general-purpose vision-language models with instruction tuning (2023)

\bibitem{fu2023mme}
Fu, C., Chen, P., Shen, Y., Qin, Y., Zhang, M., Lin, X., Yang, J., Zheng, X., Li, K., Sun, X., et~al.: Mme: A comprehensive evaluation benchmark for multimodal large language models. arXiv preprint arXiv:2306.13394  (2023)

\bibitem{ge2023chain}
Ge, J., Luo, H., Qian, S., Gan, Y., Fu, J., Zhan, S.: Chain of thought prompt tuning in vision language models. arXiv preprint arXiv:2304.07919  (2023)

\bibitem{guan2023hallusionbench}
Guan, T., Liu, F., Wu, X., Xian, R., Li, Z., Liu, X., Wang, X., Chen, L., Huang, F., Yacoob, Y., et~al.: Hallusionbench: an advanced diagnostic suite for entangled language hallucination and visual illusion in large vision-language models. In: Proceedings of the IEEE/CVF Conference on Computer Vision and Pattern Recognition. pp. 14375--14385 (2024)

\bibitem{gunjal2023detecting}
Gunjal, A., Yin, J., Bas, E.: Detecting and preventing hallucinations in large vision language models. arXiv preprint arXiv:2308.06394  (2023)

\bibitem{han2023chartllama}
Han, Y., Zhang, C., Chen, X., Yang, X., Wang, Z., Yu, G., Fu, B., Zhang, H.: Chartllama: A multimodal llm for chart understanding and generation. arXiv preprint arXiv:2311.16483  (2023)

\bibitem{hu2021lora}
Hu, E.J., Shen, Y., Wallis, P., Allen-Zhu, Z., Li, Y., Wang, S., Wang, L., Chen, W.: Lora: Low-rank adaptation of large language models. arXiv preprint arXiv:2106.09685  (2021)

\bibitem{hu2023ciem}
Hu, H., Zhang, J., Zhao, M., Sun, Z.: Ciem: Contrastive instruction evaluation method for better instruction tuning. arXiv preprint arXiv:2309.02301  (2023)

\bibitem{huang2023sparkles}
Huang, Y., Meng, Z., Liu, F., Su, Y., Collier, N., Lu, Y.: Sparkles: Unlocking chats across multiple images for multimodal instruction-following models. arXiv preprint arXiv:2308.16463  (2023)

\bibitem{hudson2019gqa}
Hudson, D.A., Manning, C.D.: Gqa: A new dataset for real-world visual reasoning and compositional question answering. In: Proceedings of the IEEE/CVF conference on computer vision and pattern recognition. pp. 6700--6709 (2019)

\bibitem{jain2024vcoder}
Jain, J., Yang, J., Shi, H.: Vcoder: Versatile vision encoders for multimodal large language models. In: Proceedings of the IEEE/CVF Conference on Computer Vision and Pattern Recognition. pp. 27992--28002 (2024)

\bibitem{jiang2023hallucination}
Jiang, C., Xu, H., Dong, M., Chen, J., Ye, W., Yan, M., Ye, Q., Zhang, J., Huang, F., Zhang, S.: Hallucination augmented contrastive learning for multimodal large language model. arXiv preprint arXiv:2312.06968  (2023)

\bibitem{krishna2017visual}
Krishna, R., Zhu, Y., Groth, O., Johnson, J., Hata, K., Kravitz, J., Chen, S., Kalantidis, Y., Li, L.J., Shamma, D.A., et~al.: Visual genome: Connecting language and vision using crowdsourced dense image annotations. International journal of computer vision  \textbf{123},  32--73 (2017)

\bibitem{lee2024meteor}
Lee, B.K., Kim, C.W., Park, B., Ro, Y.M.: Meteor: Mamba-based traversal of rationale for large language and vision models. arXiv preprint arXiv:2405.15574  (2024)

\bibitem{leng2023mitigating}
Leng, S., Zhang, H., Chen, G., Li, X., Lu, S., Miao, C., Bing, L.: Mitigating object hallucinations in large vision-language models through visual contrastive decoding. arXiv preprint arXiv:2311.16922  (2023)

\bibitem{leng2024mitigating}
Leng, S., Zhang, H., Chen, G., Li, X., Lu, S., Miao, C., Bing, L.: Mitigating object hallucinations in large vision-language models through visual contrastive decoding. In: Proceedings of the IEEE/CVF Conference on Computer Vision and Pattern Recognition. pp. 13872--13882 (2024)

\bibitem{li2023blip}
Li, J., Li, D., Savarese, S., Hoi, S.: Blip-2: Bootstrapping language-image pre-training with frozen image encoders and large language models. arXiv preprint arXiv:2301.12597  (2023)

\bibitem{li2023evaluating}
Li, Y., Du, Y., Zhou, K., Wang, J., Zhao, W.X., Wen, J.R.: Evaluating object hallucination in large vision-language models. arXiv preprint arXiv:2305.10355  (2023)

\bibitem{lightman2023let}
Lightman, H., Kosaraju, V., Burda, Y., Edwards, H., Baker, B., Lee, T., Leike, J., Schulman, J., Sutskever, I., Cobbe, K.: Let's verify step by step. arXiv preprint arXiv:2305.20050  (2023)

\bibitem{lin2024moe}
Lin, B., Tang, Z., Ye, Y., Cui, J., Zhu, B., Jin, P., Zhang, J., Ning, M., Yuan, L.: Moe-llava: Mixture of experts for large vision-language models. arXiv preprint arXiv:2401.15947  (2024)

\bibitem{lin2014microsoft}
Lin, T.Y., Maire, M., Belongie, S., Hays, J., Perona, P., Ramanan, D., Doll{\'a}r, P., Zitnick, C.L.: Microsoft coco: Common objects in context. In: Computer Vision--ECCV 2014: 13th European Conference, Zurich, Switzerland, September 6-12, 2014, Proceedings, Part V 13. pp. 740--755. Springer (2014)

\bibitem{liu2023mitigating}
Liu, F., Lin, K., Li, L., Wang, J., Yacoob, Y., Wang, L.: Mitigating hallucination in large multi-modal models via robust instruction tuning. In: The Twelfth International Conference on Learning Representations (2023)

\bibitem{liu2023improved}
Liu, H., Li, C., Li, Y., Lee, Y.J.: Improved baselines with visual instruction tuning. arXiv preprint arXiv:2310.03744  (2023)

\bibitem{liu2024visual}
Liu, H., Li, C., Wu, Q., Lee, Y.J.: Visual instruction tuning. Advances in neural information processing systems  \textbf{36} (2024)

\bibitem{liu2023mmbench}
Liu, Y., Duan, H., Zhang, Y., Li, B., Zhang, S., Zhao, W., Yuan, Y., Wang, J., He, C., Liu, Z., et~al.: Mmbench: Is your multi-modal model an all-around player? arXiv preprint arXiv:2307.06281  (2023)

\bibitem{lu2022learn}
Lu, P., Mishra, S., Xia, T., Qiu, L., Chang, K.W., Zhu, S.C., Tafjord, O., Clark, P., Kalyan, A.: Learn to explain: Multimodal reasoning via thought chains for science question answering. Advances in Neural Information Processing Systems  \textbf{35},  2507--2521 (2022)

\bibitem{magister2022teaching}
Magister, L.C., Mallinson, J., Adamek, J., Malmi, E., Severyn, A.: Teaching small language models to reason. arXiv preprint arXiv:2212.08410  (2022)

\bibitem{ouyang2022training}
Ouyang, L., Wu, J., Jiang, X., Almeida, D., Wainwright, C., Mishkin, P., Zhang, C., Agarwal, S., Slama, K., Ray, A., et~al.: Training language models to follow instructions with human feedback. Advances in Neural Information Processing Systems  \textbf{35},  27730--27744 (2022)

\bibitem{radford2021learning}
Radford, A., Kim, J.W., Hallacy, C., Ramesh, A., Goh, G., Agarwal, S., Sastry, G., Askell, A., Mishkin, P., Clark, J., et~al.: Learning transferable visual models from natural language supervision. In: International conference on machine learning. pp. 8748--8763. PMLR (2021)

\bibitem{raffel2020exploring}
Raffel, C., Shazeer, N., Roberts, A., Lee, K., Narang, S., Matena, M., Zhou, Y., Li, W., Liu, P.J.: Exploring the limits of transfer learning with a unified text-to-text transformer. The Journal of Machine Learning Research  \textbf{21}(1),  5485--5551 (2020)

\bibitem{hanoona2024LLaVA++}
Rasheed, H., Maaz, M., Khan, S., Khan, F.S.: Llava++: Extending visual capabilities with llama-3 and phi-3 (2024), \url{https://github.com/mbzuai-oryx/LLaVA-pp}

\bibitem{schuhmann2022laion}
Schuhmann, C., Beaumont, R., Vencu, R., Gordon, C., Wightman, R., Cherti, M., Coombes, T., Katta, A., Mullis, C., Wortsman, M., et~al.: Laion-5b: An open large-scale dataset for training next generation image-text models. Advances in Neural Information Processing Systems  \textbf{35},  25278--25294 (2022)

\bibitem{schunk1998self}
Schunk, D.H., Zimmerman, B.J.: Self-regulated learning: From teaching to self-reflective practice. Guilford Press (1998)

\bibitem{schwenk2022okvqa}
Schwenk, D., Khandelwal, A., Clark, C., Marino, K., Mottaghi, R.: A-okvqa: A benchmark for visual question answering using world knowledge. In: European Conference on Computer Vision. pp. 146--162. Springer (2022)

\bibitem{sun2023aligning}
Sun, Z., Shen, S., Cao, S., Liu, H., Li, C., Shen, Y., Gan, C., Gui, L.Y., Wang, Y.X., Yang, Y., et~al.: Aligning large multimodal models with factually augmented rlhf. arXiv preprint arXiv:2309.14525  (2023)

\bibitem{team2023gemini}
Team, G., Anil, R., Borgeaud, S., Wu, Y., Alayrac, J.B., Yu, J., Soricut, R., Schalkwyk, J., Dai, A.M., Hauth, A., et~al.: Gemini: a family of highly capable multimodal models. arXiv preprint arXiv:2312.11805  (2023)

\bibitem{touvron2023llama}
Touvron, H., Lavril, T., Izacard, G., Martinet, X., Lachaux, M.A., Lacroix, T., Rozière, B., Goyal, N., Hambro, E., Azhar, F., Rodriguez, A., Joulin, A., Grave, E., Lample, G.: Llama: Open and efficient foundation language models (2023)

\bibitem{touvron2023llama2}
Touvron, H., Martin, L., Stone, K., Albert, P., Almahairi, A., Babaei, Y., Bashlykov, N., Batra, S., Bhargava, P., Bhosale, S., et~al.: Llama 2: Open foundation and fine-tuned chat models. arXiv preprint arXiv:2307.09288  (2023)

\bibitem{wang2023see}
Wang, J., Meng, L., Weng, Z., He, B., Wu, Z., Jiang, Y.G.: To see is to believe: Prompting gpt-4v for better visual instruction tuning. arXiv preprint arXiv:2311.07574  (2023)

\bibitem{wang2024mitigating}
Wang, X., Pan, J., Ding, L., Biemann, C.: Mitigating hallucinations in large vision-language models with instruction contrastive decoding. arXiv preprint arXiv:2403.18715  (2024)

\bibitem{wang2022self}
Wang, X., Wei, J., Schuurmans, D., Le, Q., Chi, E., Narang, S., Chowdhery, A., Zhou, D.: Self-consistency improves chain of thought reasoning in language models. arXiv preprint arXiv:2203.11171  (2022)

\bibitem{wang2022rationale}
Wang, X., Wei, J., Schuurmans, D., Le, Q., Chi, E., Zhou, D.: Rationale-augmented ensembles in language models. arXiv preprint arXiv:2207.00747  (2022)

\bibitem{wei2022chain}
Wei, J., Wang, X., Schuurmans, D., Bosma, M., Xia, F., Chi, E., Le, Q.V., Zhou, D., et~al.: Chain-of-thought prompting elicits reasoning in large language models. Advances in Neural Information Processing Systems  \textbf{35},  24824--24837 (2022)

\bibitem{wu2023role}
Wu, Y., Zhang, P., Xiong, W., Oguz, B., Gee, J.C., Nie, Y.: The role of chain-of-thought in complex vision-language reasoning task. arXiv preprint arXiv:2311.09193  (2023)

\bibitem{ye2023mplug}
Ye, Q., Xu, H., Xu, G., Ye, J., Yan, M., Zhou, Y., Wang, J., Hu, A., Shi, P., Shi, Y., et~al.: mplug-owl: Modularization empowers large language models with multimodality. arXiv preprint arXiv:2304.14178  (2023)

\bibitem{yin2023woodpecker}
Yin, S., Fu, C., Zhao, S., Xu, T., Wang, H., Sui, D., Shen, Y., Li, K., Sun, X., Chen, E.: Woodpecker: Hallucination correction for multimodal large language models. arXiv preprint arXiv:2310.16045  (2023)

\bibitem{yin2024lamm}
Yin, Z., Wang, J., Cao, J., Shi, Z., Liu, D., Li, M., Huang, X., Wang, Z., Sheng, L., Bai, L., et~al.: Lamm: Language-assisted multi-modal instruction-tuning dataset, framework, and benchmark. Advances in Neural Information Processing Systems  \textbf{36} (2024)

\bibitem{yu2023hallucidoctor}
Yu, Q., Li, J., Wei, L., Pang, L., Ye, W., Qin, B., Tang, S., Tian, Q., Zhuang, Y.: Hallucidoctor: Mitigating hallucinatory toxicity in visual instruction data (2023)

\bibitem{yu2023mm}
Yu, W., Yang, Z., Li, L., Wang, J., Lin, K., Liu, Z., Wang, X., Wang, L.: Mm-vet: Evaluating large multimodal models for integrated capabilities. arXiv preprint arXiv:2308.02490  (2023)

\bibitem{zelikman2022star}
Zelikman, E., Wu, Y., Mu, J., Goodman, N.: Star: Bootstrapping reasoning with reasoning. Advances in Neural Information Processing Systems  \textbf{35},  15476--15488 (2022)

\bibitem{zhai2023halle}
Zhai, B., Yang, S., Zhao, X., Xu, C., Shen, S., Zhao, D., Keutzer, K., Li, M., Yan, T., Fan, X.: Halle-switch: Rethinking and controlling object existence hallucinations in large vision language models for detailed caption. arXiv preprint arXiv:2310.01779  (2023)

\bibitem{internlmxcomposer}
Zhang, P., Dong, X., Wang, B., Cao, Y., Xu, C., Ouyang, L., Zhao, Z., Ding, S., Zhang, S., Duan, H., Zhang, W., Yan, H., Zhang, X., Li, W., Li, J., Chen, K., He, C., Zhang, X., Qiao, Y., Lin, D., Wang, J.: Internlm-xcomposer: A vision-language large model for advanced text-image comprehension and composition. arXiv preprint arXiv:2309.15112  (2023)

\bibitem{zhang2022opt}
Zhang, S., Roller, S., Goyal, N., Artetxe, M., Chen, M., Chen, S., Dewan, C., Diab, M., Li, X., Lin, X.V., et~al.: Opt: Open pre-trained transformer language models. arXiv preprint arXiv:2205.01068  (2022)

\bibitem{zhang2023llavar}
Zhang, Y., Zhang, R., Gu, J., Zhou, Y., Lipka, N., Yang, D., Sun, T.: Llavar: Enhanced visual instruction tuning for text-rich image understanding. arXiv preprint arXiv:2306.17107  (2023)

\bibitem{zhang2023multimodal}
Zhang, Z., Zhang, A., Li, M., Zhao, H., Karypis, G., Smola, A.: Multimodal chain-of-thought reasoning in language models. arXiv preprint arXiv:2302.00923  (2023)

\bibitem{zhao2023svit}
Zhao, B., Wu, B., Huang, T.: Svit: Scaling up visual instruction tuning. arXiv preprint arXiv:2307.04087  (2023)

\bibitem{zhao2023chatspot}
Zhao, L., Yu, E., Ge, Z., Yang, J., Wei, H., Zhou, H., Sun, J., Peng, Y., Dong, R., Han, C., et~al.: Chatspot: Bootstrapping multimodal llms via precise referring instruction tuning. arXiv preprint arXiv:2307.09474  (2023)

\bibitem{zheng2023judging}
Zheng, L., Chiang, W.L., Sheng, Y., Zhuang, S., Wu, Z., Zhuang, Y., Lin, Z., Li, Z., Li, D., Xing, E.P., Zhang, H., Gonzalez, J.E., Stoica, I.: Judging llm-as-a-judge with mt-bench and chatbot arena (2023)

\bibitem{zhu2023minigpt}
Zhu, D., Chen, J., Shen, X., Li, X., Elhoseiny, M.: Minigpt-4: Enhancing vision-language understanding with advanced large language models. arXiv preprint arXiv:2304.10592  (2023)

\bibitem{zong2024mova}
Zong, Z., Ma, B., Shen, D., Song, G., Shao, H., Jiang, D., Li, H., Liu, Y.: Mova: Adapting mixture of vision experts to multimodal context. arXiv preprint arXiv:2404.13046  (2024)

\end{thebibliography}

\clearpage
\appendix
\renewcommand\thesection{\Alph{section}}
\renewcommand\thefigure{A\arabic{figure}}
\renewcommand\thetable{\Roman{table}}
\definecolor{Gray}{gray}{0.93}
\title{Reflective Instruction Tuning: Mitigating Hallucinations in Large Vision-Language Models} 

\titlerunning{Reflective Instruction Tuning}

\author{\Large{Supplementary Materials}}

\authorrunning{J.~Zhang et al.}

\institute{\large{\url{https://zjr2000.github.io/projects/reverie}}
}

\maketitle

\section{Prompts for Dataset Annotation}

\begin{figure}[]
  \centering
  \includegraphics[width=\textwidth]{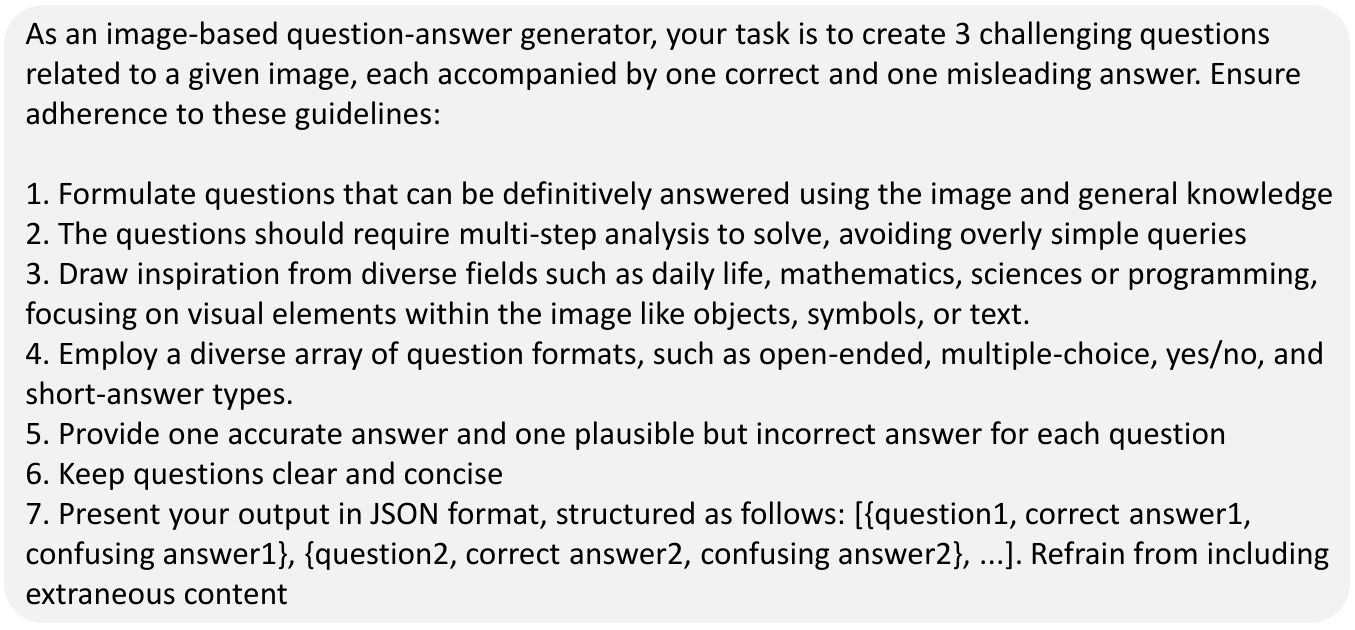}
  \caption{\textbf{Prompt for instruction-response generation}} 
  \label{fig:prompt_qa}
\end{figure}

\begin{figure}[]
  \centering
  \includegraphics[width=\textwidth]{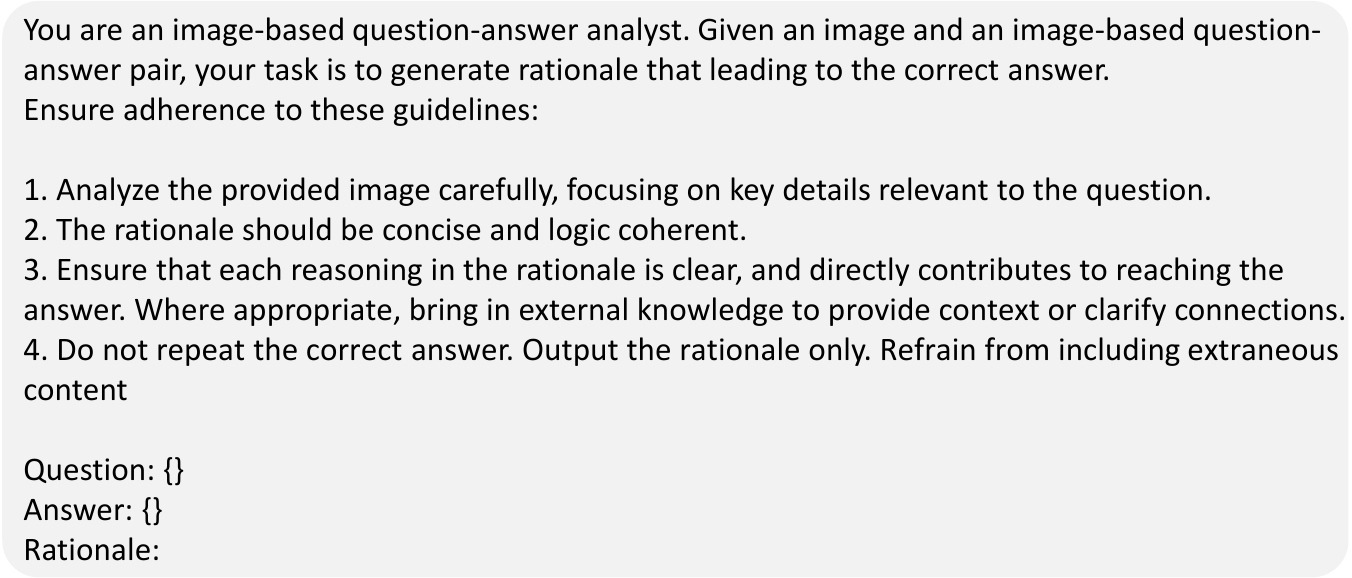}
  \caption{\textbf{Prompt for positive rationale generation}} 
  \label{fig:prompt_pos}
\end{figure}

\begin{figure}[]
  \centering
  \includegraphics[width=0.9\textwidth]{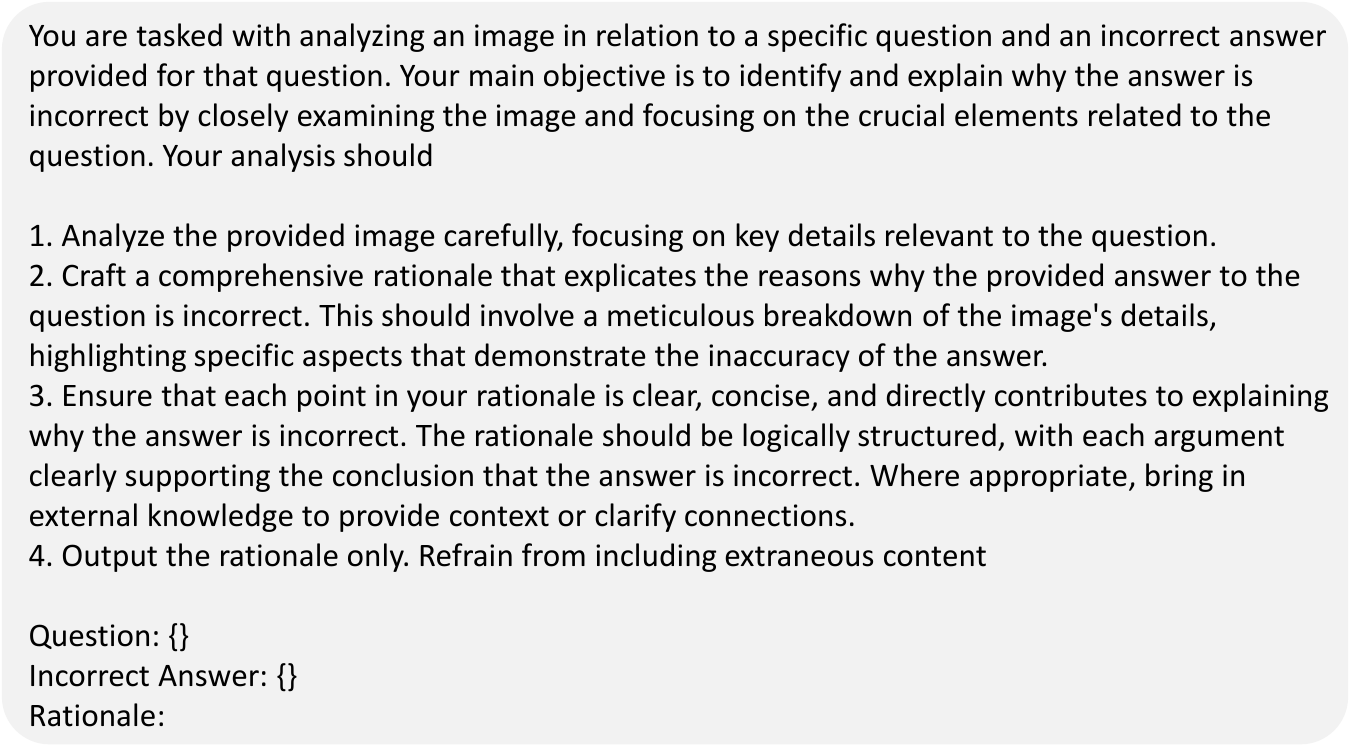}
  \caption{\textbf{Prompt for negative rationale generation}} 
  \label{fig:prompt_neg}
\end{figure}

\begin{figure}[]
  \centering
  \includegraphics[width=\textwidth]{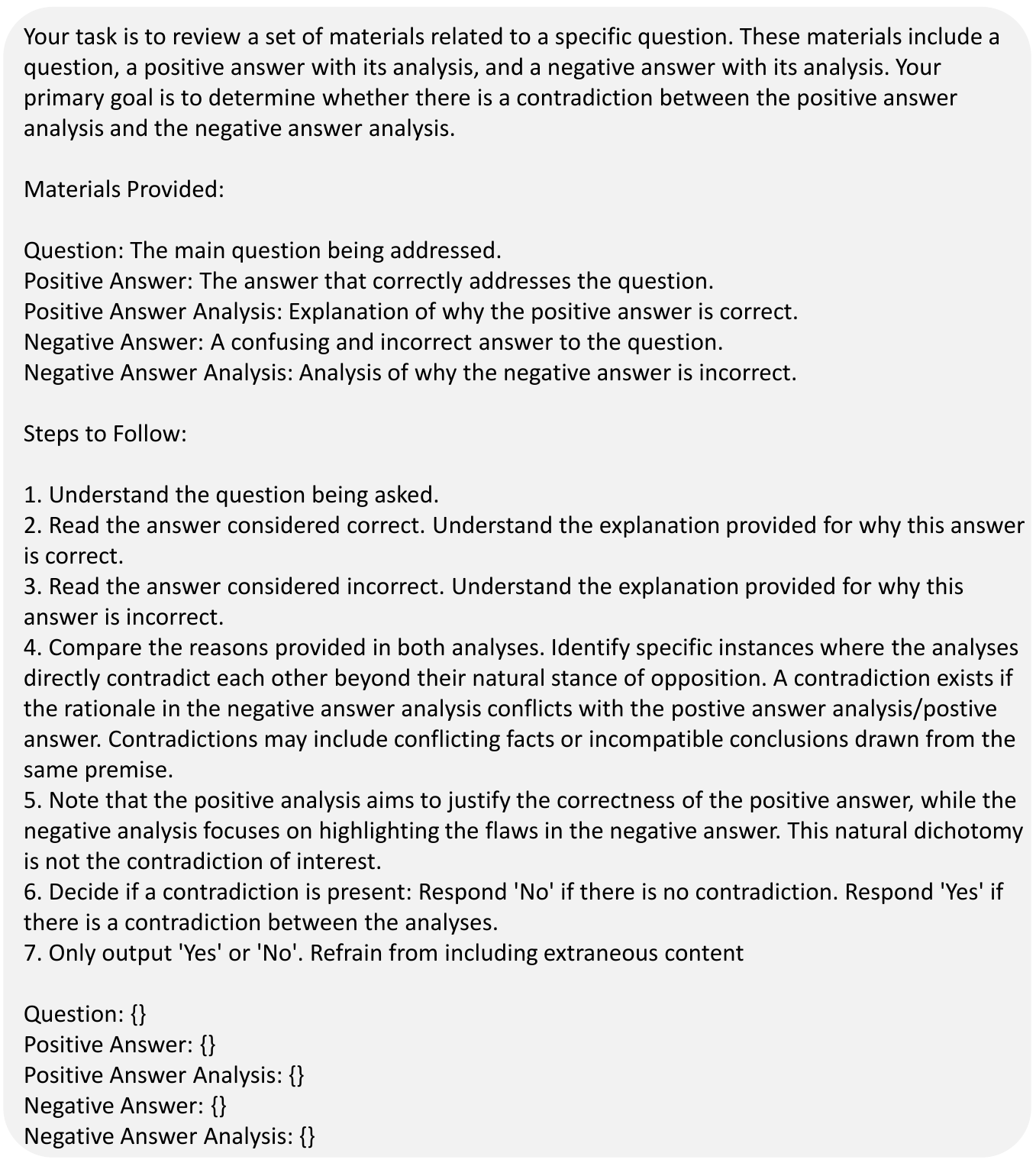}
  \caption{\textbf{Prompt for consistency-based data filtering}} 
  \label{fig:prompt_check}
\end{figure}

This section presents the prompts employed for annotating the \textit{REVERIE} dataset. Fig.~\ref{fig:prompt_qa} shows the prompt for the generation of instruction and response, we ask Gemini~\cite{team2023gemini} vision model to generate instructions that require multi-step reasoning, span diverse domains and encompass various task types. Additionally, for each instruction, we prompt the model to generate both a positive and a negative response.  Fig.~\ref{fig:prompt_pos} and Fig~\ref{fig:prompt_neg} present the prompt for positive rationale and negative rationale generation, respectively. We instruct the model to show the reasoning underlying answer prediction and also include key image details and related outside knowledge in the rationale. Fig.~\ref{fig:prompt_check} shows the prompt for consistency-based data filtering. We instruct the model to check whether there is a contradiction between positive and negative rationale (\eg, conflict facts).

\begin{table}[t]
\captionof{table}{\textbf{Variants of rationale generation prompts}.}
\centering
\scalebox{0.8}{
\begin{tabular}{l |m{69mm}| m{68mm}}
\toprule
\textbf{Line} & \textbf{Positive rationale generation prompt} & \textbf{Negative rationale generation prompt} \\
\midrule
(a) & {\small Explain why.} &  {\small Explain why this answer is wrong : \{incorrect answer\}.} \\
\midrule
(b) & {\small Explain why this answer is correct. Analyze the image, identifying key details and explaining how they lead to the answer. If necessary, use external knowledge to clarify your reasoning.} &  {\small Explain why this answer is wrong : \{incorrect answer\}.Analysis the image, highlighting specific details that demonstrate the inaccuracy of the answer. If necessary, use external knowledge to clarify your reasoning.} \\
\midrule
(c) & {\small 1. Explain why this answer is correct in one to two simple sentences, including any necessary facts or knowledge.\newline 2. Explain why this answer is correct, including any necessary facts or knowledge.\newline 3. Explain why this answer is correct. Analyze the image, identifying key details and explaining how they lead to the answer. If necessary, use external knowledge to clarify your reasoning.} & {\small Explain why this answer is wrong : \{incorrect answer\}. Analysis the image, highlighting specific details that demonstrate the inaccuracy of the answer. If necessary, use external knowledge to clarify your reasoning.} \\
\midrule
(d) & {\small Explain why. Including any necessary facts or knowledge} & {\small Explain why this answer is wrong: \{incorrect answer\}. Including any necessary facts or knowledge.} \\

\bottomrule
\end{tabular}
}
\label{tab:rationale_prompts}
\end{table}

\section{Implementation Details}

\noindent{\textbf{Evaluation Benchmarks}} We perform evaluation on 6 benchmarks: POPE~\cite{li2023evaluating}, which assesses object hallucination in LVLMs by querying object existence on the COCO dataset; GQA~\cite{hudson2019gqa}, a visual question answering benchmark featuring open-ended questions; ScienceQA~\cite{lu2022learn}, a multiple-choice question benchmark focused on scientific domains; MMBench~\cite{liu2023mmbench}, a comprehensive LVLM evaluation benchmark with various multiple-choice questions; MME~\cite{fu2023mme}, a comprehensive LVLM evaluation benchmark with diverse Yes/No questions; MM-Vet~\cite{yu2023mm}, a benchmark designed to examine LVLMs on complex multimodal tasks.

\noindent{\textbf{Training.}} We follow the same training recipe as the official implementations~\cite{liu2024visual} of LLaVA-1.0-7b-lora and LLaVA-1.5-7b-lora to train our models. For REVERIE-1.0-7b-lora, we select the model with CLIP-ViT-Large-336px~\cite{radford2021learning} as encoder and Vicuna-1.3-7b~\cite{zheng2023judging} as LLM, and using LoRA~\cite{hu2021lora} to train the LLM and linear projector adhering to lighting training instructions from the LLaVA's codebase. For REVERIE-1.5-7b-lora, we train the MLP connector and LLM with LoRA. Both REVERIE-1.0-7b-lora and REVERIE-1.5-7b-lora employ a linear warmup for the first 3\% of training steps followed by cosine decay scheduling.  We use a learning rate of 2e-5 for REVERIE-1.0-7b-lora. For REVERIE-1.5-7b-lora, the MLP connector learning rate is 2e-5, while all other parameters use a learning rate of 2e-4.  Both models use a global batch size of 128, train for one epoch, and utilize a LoRA attention dimension of 128 with an alpha of 256.

\noindent{\textbf{Ablation of rationale generation prompt.}} Table 6 shows the results with different rationale generation prompts. The prompts used for each variant are shown in Table~\ref{tab:rationale_prompts}. Line (a) shows the prompt that contains only a simple query. Line (d) shows our final rationale generation prompt that only includes explicit guidance. Line (b) shows the type-specific prompt. The positive rationale generation prompt emphasizes the reasoning from visual or outside evidence to answers, while the negative rationale generation prompt emphasizes the crucial details that indicate the inaccuracy of the answers. Line (c) additionally adds prompts for length control. Specifically, the positive rationale from ScienceQA~\cite{lu2022learn} and A-OK-VQA~\cite{schwenk2022okvqa} are shorter than the rationale generated from Gemini. The rationales from A-OK-VQA are concise (1-2 sentences), and the rationale from ScienceQA are shorter because they usually omit description to image details. Thus, we design specific prompts (\ie, the first two prompts in Line(c)) that are more suitable for their lengths.

\begin{table}[]
    \centering
    \caption{\textbf{Performance comparison on more LVLMs.}}
    \scalebox{0.94}{
    \begin{tabular}{@{}l|c|ccc|c@{}}
        \toprule
        & Date & POPE & MMBench & ScienceQA & Average \\ 
        \midrule
        MOE-LLaVA-1.6Bx4~\cite{lin2024moe} & 24/01 & 85.9 & 63.3 & 63.9 & 71.0 \\
        \rowcolor{Gray}
        MOE-REVERIE-1.6Bx4 & - & \textbf{86.7} & \textbf{64.5} & \textbf{77.1} & \textbf{76.1} \\
        \midrule
        LLaVA-Phi3-LoRA~\cite{hanoona2024LLaVA++} & 24/04 & 85.6 & 68.2 & 73.8 & 76.2 \\
        \rowcolor{Gray}
        REVERIE-Phi3-LoRA & - & \textbf{86.3} & \textbf{69.0} & \textbf{86.7} & \textbf{80.7} \\ 
        \bottomrule
    \end{tabular}}
    \label{tab:stronger_models}
\end{table}

\begin{table}[]
    \centering
    \caption{\textbf{Results on MMHal-Bench}. We use gpt-4-turbo-2024-04-09 as the evaluator. The term ``lora" has been omitted to save space.}
    \scalebox{0.88}{
    \begin{tabular}{@{}l|cc|cc@{}}
        \toprule
        & LLaVA-1.0-7b & REVERIE-1.0-7b & LLaVA-1.5-7b & REVERIE-1.5-7b \\ 
        \midrule
        \textbf{Avg. Score($\uparrow$)} & 1.39 & \textbf{1.43} \textcolor{black}{$(+2.9\%)$} & 2.23 & \textbf{2.36} \textcolor{black}{$(+5.8\%)$} \\
        \textbf{Halluc. Rate ($\downarrow$)} & 0.76 & \textbf{0.73} \textcolor{black}{$(-3.9\%)$} & 0.55 & \textbf{0.50} \textcolor{black}{$(-9.1\%)$} \\ \bottomrule
    \end{tabular}
    }
    \label{tab:mmhal}
\end{table}

\section{Additional Results}
\noindent\textbf{Results on additional LVLMs.} To further justify the effectiveness of reflective instruction tuning, we conduct experiments on two additional models: MoE-LLaVA~\cite{lin2024moe}, employing a mixture-of-experts design, and LLaVA-Phi3~\cite{hanoona2024LLaVA++}, incorporating a stronger LLM. As shown in Table~\ref{tab:stronger_models}, two stronger baselines with reflective tuning using our data show consistently improved performance.

\noindent\textbf{Results on MMHal-Bench.} We conduct experiments on the MMHal-Bench~\cite{sun2023aligning} to further show the effectiveness of reflective instruction tuning on hallucination mitigation. As shown in Table~\ref{tab:mmhal}, ours consistently outperforms the baselines in terms of both hallucination rate and average score.

\section{Visualizations}
\begin{figure}[]
  \centering
  \includegraphics[width=\textwidth]{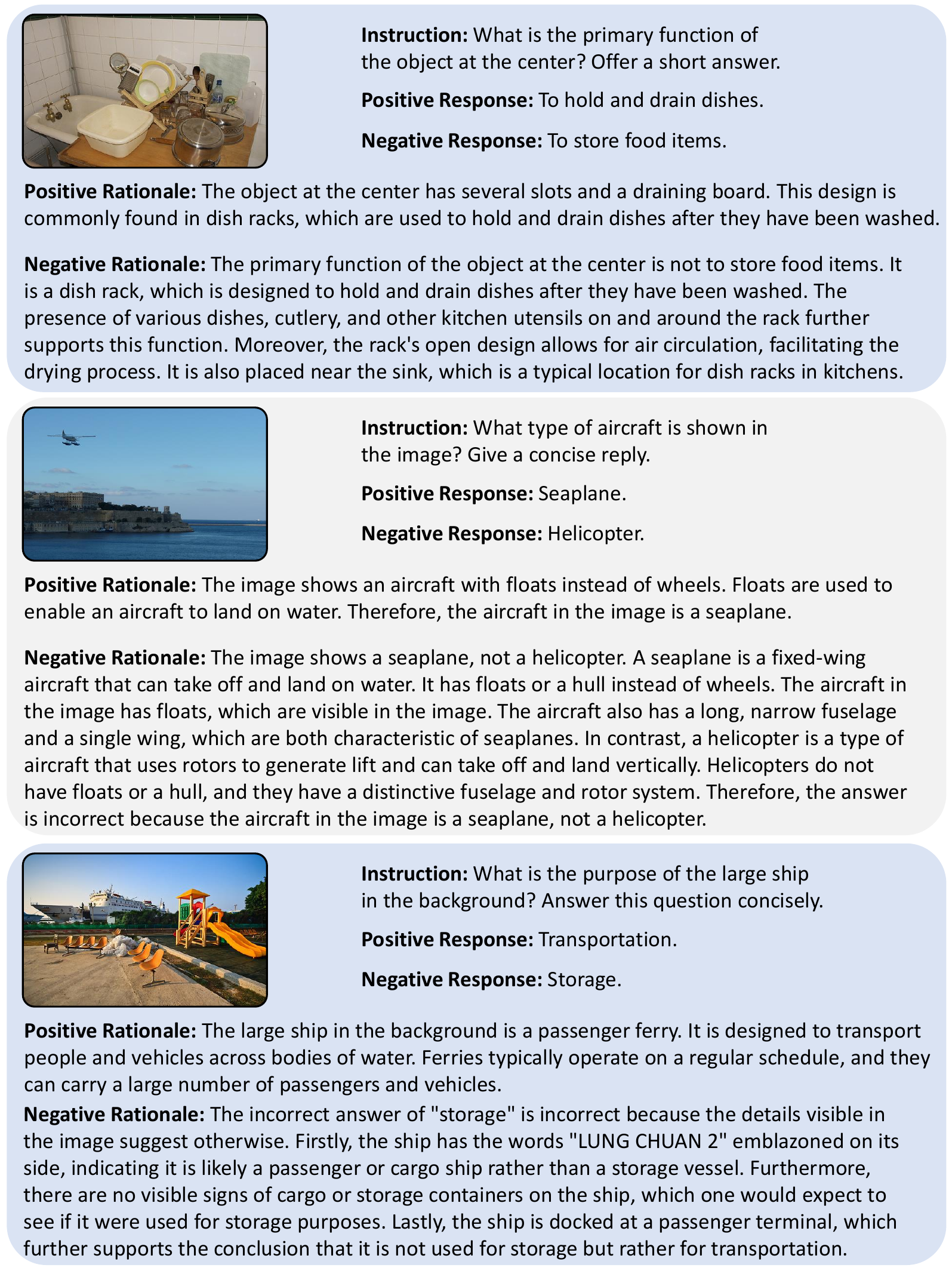}
  \caption{\textbf{Examples of the \textit{REVERIE} dataset}} 
  \label{fig:example1}
\end{figure}

\begin{figure}[]
  \centering
  \includegraphics[width=\textwidth]{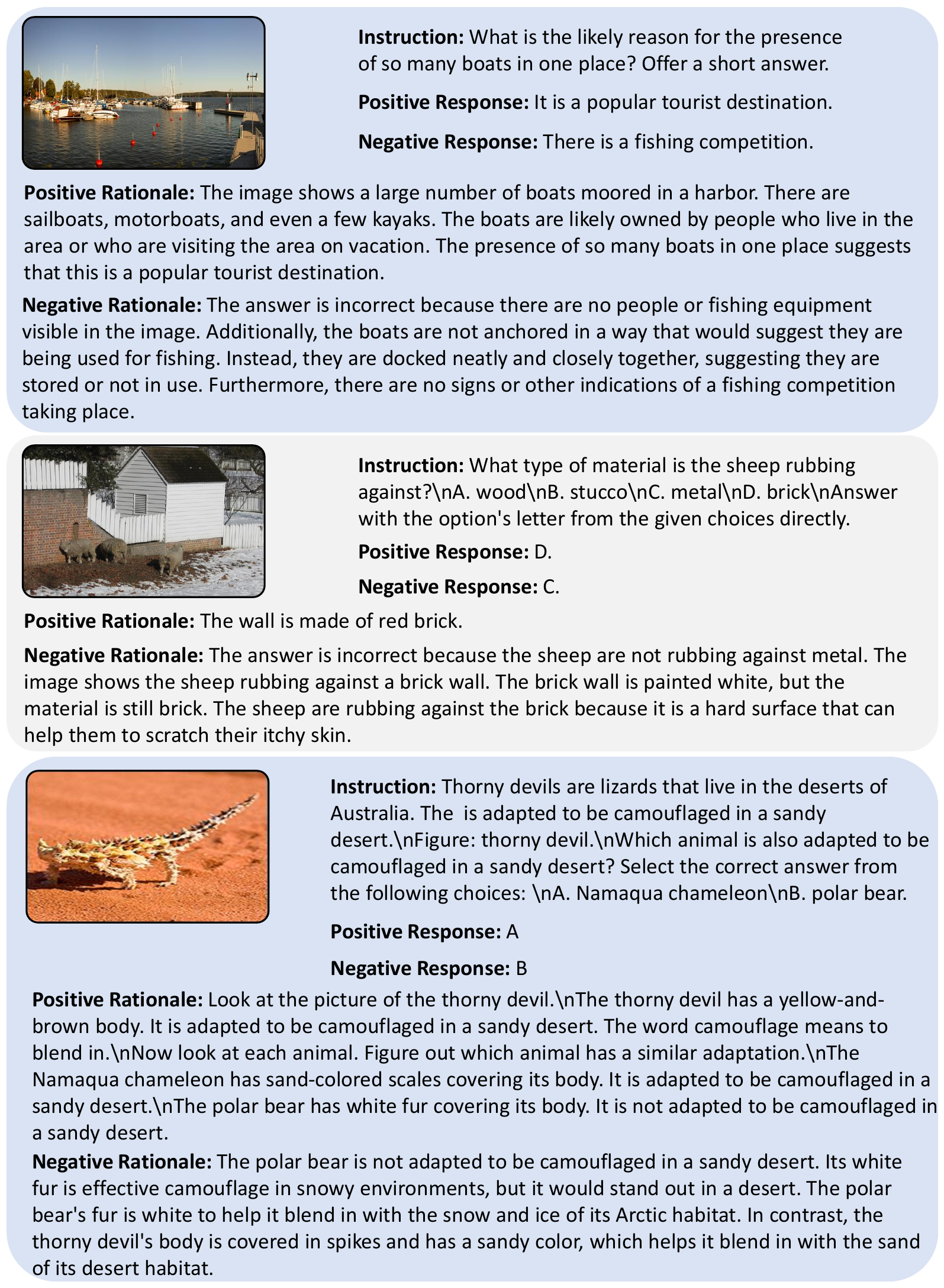}
  \caption{\textbf{Examples of the \textit{REVERIE} dataset}} 
  \label{fig:example2}
\end{figure}

Fig.~\ref{fig:example1} and Fig.~\ref{fig:example2} present examples from the \textit{REVERIE} dataset. The rationales contain rich visual information, outside knowledge and underlying logic, providing fine-grained reasoning supervision that helps address hallucinations.

\end{document}